\documentclass[runningheads]{llncs}
\usepackage{graphicx}
\usepackage{subfigure}
\usepackage{amsmath}
\usepackage{xcolor}
\usepackage{microtype}
\usepackage{fancyhdr}
\begin{document}
\title{A New Image Quality Database for Multiple Industrial Processes}
\author{\textls[-20]{Xuanchao Ma\inst{1\textrm{-}5}, Yanlin Jiang\inst{1\textrm{-}5}, Hongyan Liu\inst{1\textrm{-}5}, Chengxu Zhou \inst{1\textrm{-}7}, Ke Gu\inst{1\textrm{-}5}}}

\institute{
Faculty of Information Technology, Beijing University of Technology, Beijing, China\and
\textls[-85]{Engineering Research Center of Intelligent Perception and Autonomous Control, Ministry of Education}\and
Beijing Laboratory of Smart Environmental Protection\and
Beijing Key Laboratory of Computational Intelligence and Intelligent System\and
Beijing Artificial Intelligence Institute\and
\textls[-80]{School of Electronic \& Information Engineering, Liaoning University of Technology, Liaoning, China}\and
{Key Laboratory of Intelligent Control and Optimization for Industrial Equipment of \\ Ministry of Education, Dalian University of Technology, China}\\
Contacting Email:  zhouchengxu@lnut.edu.cn}

\maketitle
\begin{abstract}
Recent years have witnessed a broader range of applications of image processing technologies in multiple industrial processes, such as smoke detection, security monitoring, and workpiece inspection. Different kinds of distortion types and levels must be introduced into an image during the processes of acquisition, compression, transmission, storage, and display, which might heavily degrade the image quality and thus strongly reduce the final display effect and clarity. To verify the reliability of existing image quality assessment methods, we establish a new industrial process image database (IPID), which contains 3000 distorted images generated by applying different levels of distortion types to each of the 50 source images. We conduct the subjective test on the aforementioned 3000 images to collect their subjective quality ratings in a well-suited laboratory environment. Finally, we perform comparison experiments on IPID database to investigate the performance of some objective image quality assessment algorithms. The experimental results show that the state-of-the-art image quality assessment methods have difficulty in predicting the quality of images that contain multiple distortion types.

\end{abstract}

\section{Introduction}
As a cutting-edge technology, industrial vision technology has unique advantages in the industrial field. Image-based industrial vision technology has been widely used in industrial processes, such as smoke detection [\ref{ref:1}-\ref{ref:3}], security monitoring [\ref{ref:4}], and workpiece inspection [\ref{ref:5}]. With the increase in application environment and the continuous improvement of application requirements, the requirements for industrial vision technology in industrial processes are also constantly improving, and the requirements for image quality are getting higher and higher. However, in the industrial revolution, many factors, such as noise, radiation, interference, and so on, resulting in overexposure or underexposure, low visibility, motion blur, out-of-focus, and other distortion problems, inevitably lead to image quality degradation. In addition, image acquisition, compression, transmission, and storage will inevitably lead to image quality decline in industrial processes.   Therefore, it is imperative to conduct extensive subjective and objective experiments to evaluate the perceptual quality of corrupted industrial images, further serving as a trial bed for future image enhancement technology research.

With the development of industrial vision technology, image quality assessment (IQA) has been widely concerned. The IQA method is one of the basic techniques of image processing. It can evaluate the distortion degree by analyzing the features of the image. Then it can be applied to the fields of super-resolution reconstruction [\ref{ref:6}], image repair [\ref{ref:7}], image enhancement [\ref{ref:8},\ref{ref:9}], image denoising [\ref{ref:10}], image compression [\ref{ref:11}], and image dehazing [\ref{ref:12},\ref{ref:13}]. Through the in-depth study of the IQA method, researchers have formed a relatively stable research direction and ideas. IOA methods can be divided into subjective image quality assessment and objective image quality assessment according to different assessment subjects. Subjective assessment methods rely on human observers to evaluate [\ref{ref:14},\ref{ref:15}]. In this method, subjects view images and give subjective ratings that reflect their perception of the quality of the pictures. The objective assessment method mainly uses the machine algorithm as the assessment subject to get the prediction result to reflect the subjective perception of human eyes.

Although subjective IQA methods have high accuracy, the assessment process is complex, and the results are affected by individual differences and emotional factors. The objective IQA method has the characteristics of fast speed and stable results and can be widely used in various scenarios. Therefore, many researchers are devoted to exploring practical objective IQA algorithms. Accurate image quality assessment methods can be divided into full-reference image quality assessment (FR-IQA) [\ref{ref:16},\ref{ref:17}], reduced-reference image quality assessment (RR-IQA) [\ref{ref:18},\ref{ref:19}], and no-reference image quality assessment (NR-IQA) [\ref{ref:20}-\ref{ref:25}] according to the degree of dependence on reference images. The FR-IQA algorithm needs all the information of the reference image; that is, it can evaluate the image quality by comparing the reference image with the distorted image pixel by pixel. RR-IQA algorithm selects some information from the reference image to assess the distorted image. Its algorithm complexity is lower than the FR-IQA algorithm but can not be widely used because it references some reference information. The NR-IQA algorithm, the image blind assessment algorithm[\ref{ref:26}-\ref{ref:29}], is the most commonly used in practical scenarios because it can effectively evaluate distorted images without reference information.

The traditional FR-IQA methods are MSE [\ref{ref:30}] and PSNR. However, due to the fact that PSNR is sometimes associated with human visual features, more and more researchers have worked hard to develop some assessment algorithms based on human perception, such as structural similarity (SSIM) [\ref{ref:31}], visual information fidelity (VIF) [\ref{ref:32}], and visual signal-to-noise ratio (VSNR) [\ref{ref:33}]. Due to the distortion caused by the external environment and internal camera sensors, it is challenging to obtain distortion-free images. Therefore, people prefer to use the NR-IQA method to evaluate the quality of distorted images. NIQMC(blind/No-reference Image Quality Metric for Contrast distortion) [\ref{ref:34}] is a non-reference image quality assessment algorithm based on image information maximization proposed by Gu et al. in 2017, which measures the overall quality of distorted images from local and global perspectives, respectively. BIQME(Blind
Image Quality Measure of Enhanced images) [\ref{ref:35}] is a reference-free image quality detection framework proposed by Gu et al. The algorithm considers five factors of image contrast, sharpness, brightness, color, and naturalness and extracts 17 features. Most NR-IQA models, such as NIQE [\ref{ref:36}], NFERM [\ref{ref:37}], ARIS [\ref{ref:38}], etc., are built on natural scene statistics (NSS) or the human Visual system (HVS).

However, much less effort has been done to assess the perceptual quality of corrupted industrial process images. To verify the reliability of existing image quality assessment methods, this paper establishes the Industrial Process Image Database (IPID), which consists of 3000 distorted images generated by applying different types and degrees of distortion to 50 source images. We conduct a subjective scoring experiment on the proposed IPID to obtain the image quality assessment score. To study the performance of some objective image quality assessment algorithms, we further carry out a comparative experiment on IPID. The experimental results show that the existing IQA algorithm has little correlation with the subjective score and can not evaluate the quality of distorted images well. It still needs to be solved to assess image quality with multiple distortion types using existing image quality assessment methods.

The rest of this paper is structured in the following way: Section 2 gives the database construction process and subjective experiment. Section 3 implements comparison tests on the IPID database to examine the performance of mainstream and state-of-the-art objective image quality assessment models. We summarize this paper in Section 4.

\section{Subjective Quality Assessment}
\subsection{Database Construction}
We construct a specific image database comprising various distorted industrial process images for assessment of image quality. When building the IPID database, we first need to collect 50 industrial process source images, and when collecting these images, we need to ensure that the images are distortion-free. In capturing source images, we choose different periods to shoot, such as day and night, to obtain different light conditions and atmospheres. We shoot in various industrial process environments, such as workshops, production lines, warehouses, etc., to show a variety of industrial process scenes. The images collected mainly include pictures of industrial process elements such as workshops, assembly lines, torches, and close-up shots to build a diversified IPID database. By constructing these 50 undistorted pictures as required into an IPID database and ensuring that the content of each image is rich and diverse, it can fully demonstrate the characteristics of industrial processes.

\begin{figure}[!t]
	\small
	\centering
	\includegraphics[width=12cm]{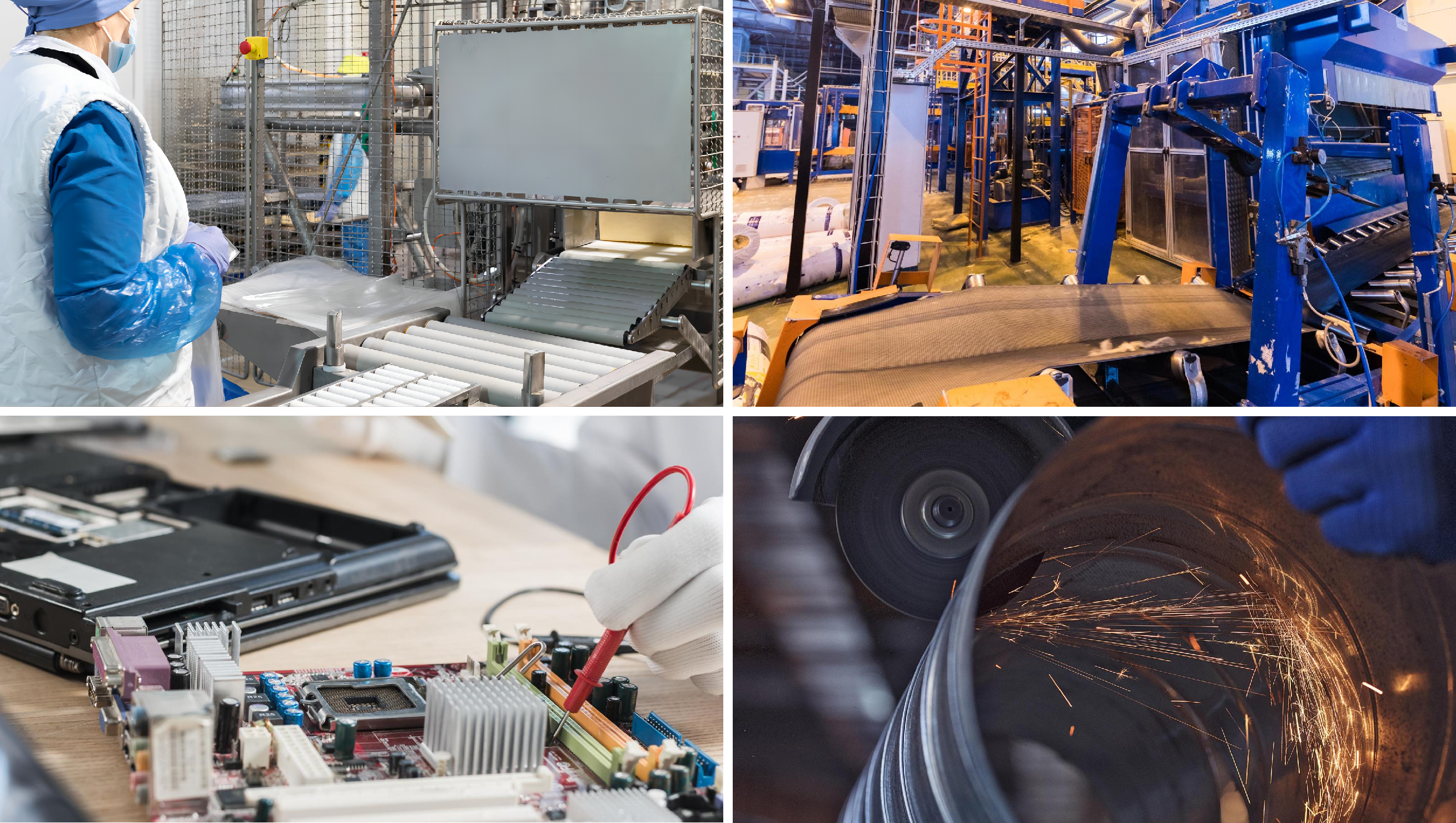}
	\caption{\small  Four sample photos in the industry that used to construct the IPID database. }
	\label{fig:2}
	\vspace{-0.25cm}
\end{figure}

 To simulate various distortion effects such as overexposure, motion blur, and out of focus that may occur during image capture, compression, transmission, and storage of industrial process images, we introduce 11 different types of distortion in 50 source images. These distortions include Gaussian blur, lens blur, motion blur, JPEG compression, Gaussian noise, H.264/AVC, H.265/HEVC, brightness changes, dark changes, average offset, and contrast changes. Among them, Gaussian blur is used to simulate problems such as lens jitter and focusing anomaly; lens blur is used to simulate lens blur that may occur in complex industrial processes; the motion blur is used to simulate the blur effect caused by the motion of the subject or camera; JPEG compression is used to reduce the information content of image data to affect the application of compression coding technology; Gaussian noise is used to simulate the image acquisition process Gaussian noise; H.264/AVC is a standard method for image compression; H.265/HEVC is another traditional method for image compression; the brightness change is used to simulate the brightness change of the industrial process image; the darkness change is used to affect the dark adaptation of the image; the average offset is used to simulate the slight shift of the position of the lens or the subject and the contrast change is used to affect the contrast change of the image. Finally, 3000 distorted images are generated from 50 original images to build an IPID database by introducing different types and degrees of distorted images. By introducing these distorting effects, we can better act on the various problems that can arise in industrial processes and provide challenging industrial process image samples in the IPID database. Such a database will help researchers and developers better understand and deal with the impact of these distorting effects on image quality and processing algorithms.

\subsection{Subjective Experiment}
In this section, we conduct a subjective experiment on the proposed IPID to collect subjective quality assessment scores of images. By collecting rating data from many participants, we can calculate each industrial process image's average subjective quality score.

Since obtaining an utterly distortion-free image as a reference image in practical applications is difficult, we choose to use the single stimulus (SS) method for experiments according to the recommendation of the International Telecommunication Union (ITU) [\ref{ref:39}]. The SS method is an experimental method used to evaluate image quality, which is suitable for the situation where many image quality scores need to be obtained quickly. In the experiment, participants receive one image as a stimulus and are asked to rate it for quality. This method is often used in image database construction and image quality assessment [\ref{ref:40}].

The fifty college students between the ages of 23 and 26 participate in our subjective assessment experiment. They all have a standard or corrected vision. Before the formal personal investigation begins, we ensure that all participants have understood the experiment's purpose and the testing process. In this experiment, the first step is a brief training course aimed at introducing participants to how to evaluate the quality of different images. Participants will learn to use the given scoring methods and standards to evaluate image quality. This can include explaining the meaning of each rating option and how to classify images based on perceived quality features. The training course will provide some sample images to help participants understand how to apply scoring methods. These sample images may represent images of different quality levels, and participants can rate them and compare them with predetermined answers. During the training process, discussions and interactions will be conducted to enable participants to raise questions, clarify doubts, and share experiences and perspectives with other participants.In the second formal testing phase, participants will use MATLAB Graphical User Interface (GUI) to rate the images, as shown in Figure 2. The GUI interface may provide participants with functions such as image display areas, rating options, and submit buttons. Participants need to rate each image based on the scoring methods and standards they have learned during the training. To avoid errors caused by visual fatigue, each participant is given a 5-minute rest time every 20 minutes of the testing process. This rest period can help their eyes relax and recover, ensuring that they can maintain focus and accuracy before the next round of testing, and maintain their visual health. Through such training and testing arrangements, we aim to ensure that participants remain focused and accurately evaluate images during the experimental process, while also protecting their visual health. This design helps to improve the accuracy and reliability of scoring, and provides participants with more support and guidance in understanding scoring methods.

We use three HP 23.8-inch monitors for all of our tests. The best resolution for these displays is 2560 x 1440, and they can be rotated to meet different resolution requirements. At the beginning of the subjective test, we set the line-of-sight to 2 or 2.5 times the screen's height. We maintain an experimental environment with moderate illumination and low noise conditions. Given the large number of images in our database, the GUI allows switching and rating of pictures via the keyboard, thus improving the efficiency of the experiment. The image quality score can be automatically saved in the form by clicking the "Submit" button. For the test images, we set five image quality ratings, namely ``poor'', ``bad'', ``fair'', ``good'' and ``excellent'', corresponding to a subjective rating of 1 to 5. A lower score indicates poor perceived quality.

\begin{figure}[!t]
	\small
	\centering
	\includegraphics[width=12cm]{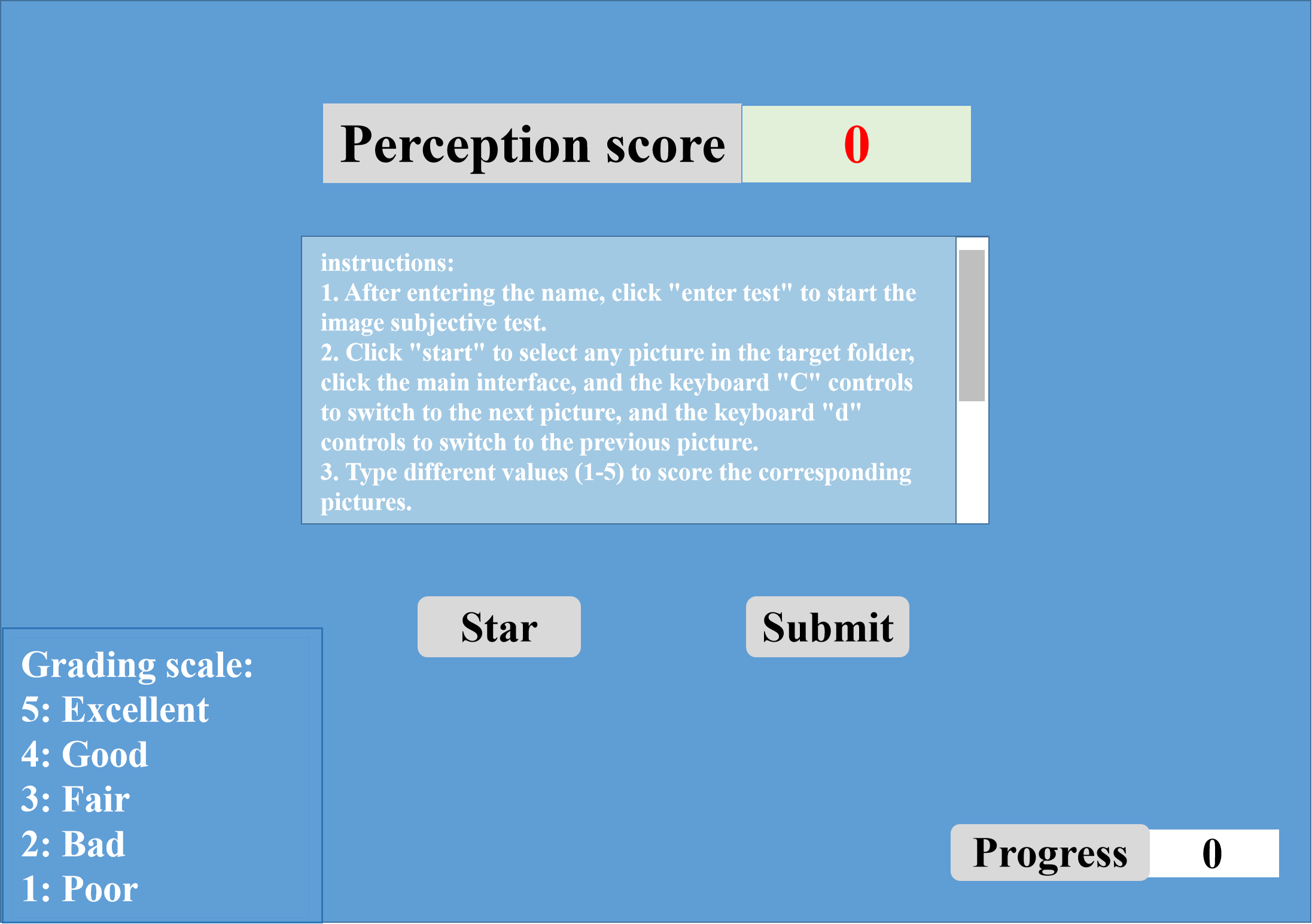}
	\caption{\small The MATLAB Graphical User Interface.}
	\label{fig:2}
	\vspace{-0.25cm}
\end{figure}

\subsection{Subjective Score Processing and Analysis}
To get a more accurate subjective score, we will remove the highest and lowest scores for each image. Then, the mean opinion score (MOS) is calculated as the final essential true value of the image quality. The MOS value for each image can be calculated using the following formula:
\begin{equation}
	MOS_j=\sum_{i=1}^{N}\frac{P_{ij}}{N}
	\label{func:1}
\end{equation}
where $P_{ij}$ represents the score of $i^{th}$ participant to $j^{th}$ image, $N$ is the number of pictures.

\section{Experiment}
In this section, we will investigate the performance of existing mainstream objective IQA methods to evaluate the perceptual quality of the images in the IPID database.
\subsection{Performance Metrics}
We utilize a five-parameter monotonic logistic function to fit the prediction results of the IQA model, so as to reduce the nonlinearity of the prediction values. The above logistic function is shown as the following:
\begin{equation}
	W(x)=\beta_1(0.5-\frac{1}{1+e^{\beta_2(x-\beta_3)}})+\beta_4x+\beta_5
	\label{func:2}
\end{equation}
where $x$ and $W$ represent the subjective scores and mapping scores, respectively.
When calculating performance indicators, we use the following five commonly used indicators to evaluate the accuracy of the prediction model: Spearman Rank-Order Correlation Coefficient (SROCC), Pearson Linear Correlation Coefficient (PLCC), Kendall Rank Correlation Coefficient (KRCC), Root Mean-Squared Error (RMSE) and Mean Absolute Error (MAE). SROCC and KRCC are used to measure the linear relationship between two vectors and evaluate the monotonicity of predictions. PLCC evaluates the linearity and consistency of the model. RMSE measures the accuracy of predictions by calculating the difference between predicted results and actual values. MAE predicts the mean absolute error. The five indices mentioned above can be expressed as follows:
\begin{equation}
 	SROCC=1-\frac{6\sum\limits_{n=1}^{N}k_i^2}{N(N^2-1)}
 	\label{func:3}
\end{equation}
\begin{equation}
	PLCC=\frac{\sum\limits_{i=1}^{N}(h_i-\overline{h})(g_i-\overline{g})}{\sqrt{\frac{1}{N}\sum\limits_{i=1}^{N}(h_i-\overline{h})^2(g_i-\overline{g}^2)}}
	\label{func:4}
\end{equation}
\begin{equation}
	KRCC=\frac{N_c-N_k}{0.5N(N-1)}
	\label{func:5}
\end{equation}
\begin{equation}
	RMSE=\sqrt{\frac{1}{N}\sum\limits_{i=1}^{N}(g_i-h_i)^2}
	\label{func:6}
\end{equation}
\begin{equation}
	MAE=\frac{1}{N}\sum\limits_{i=1}^{N}|g_i-h_i|
	\label{func:7}
\end{equation}
where $c$ and $k$ are the consistent data and the inconsistent data, respectively; $g_i$ is the subjective quality score of the $i^{th}$ image; $s$ is the prediction output of the objective quality assessment model after nonlinear mapping; $h_i$ and $g_i$ are the average values of $h$ and $g$. Among these indices, the larger the value of KRCC and PLCC, the better the predict performance; while the smaller the value of RMSE, the higher the accuracy.

\subsection{Comparison Experiment and Result Analysis}
In this section, we select 13 kinds of objective quality assessment approaches to assess the images in our established database, including SSIM, multi-SSIM (MS-SSIM) [\ref{ref:41}],  visual information fidelity in pixel domain [\ref{ref:42}], feature similarity (FSIM) [\ref{ref:43}], FSIMc, gradient similarity index (GSI) [\ref{ref:44}], gradient magnitude similarity (GMSM), gradient magnitude similarity deviation (GMSD) [\ref{ref:45}], local-tuned-global model (LTG) [\ref{ref:46}], visual saliency induced index (VSI) [\ref{ref:47}], analysis of distribution for pooling in SSIM (ADD-SSIM) [\ref{ref:48}], ADD-GSIM, and perceptual similarity (PSIM) [\ref{ref:49}]. The comparison experiment results of these methods are shown in Table 1, where the one with the best performance is marked in bold.

\begin{table}[htbp]
\caption{Performance comparison of thirteen state-of-the-art FR-IQA models on the
IPID dataset}
\centering
\renewcommand\arraystretch{1.5}{
\setlength{\tabcolsep}{4mm}{
\begin{tabular}{|c|c|c|c|c|c|}
\hline
Algorithm & PLCC & SROCC & KROCC & RMSE & MAE \\
\hline
SSIM & 0.4179 & 0.3890 & 0.2740 & 0.5800 & 0.4658\\
\hline
MS-SSIM & 0.6058 & 0.5756 & 0.3976 & 0.5080 & 0.3984 \\
\hline
VIFP & \textbf{0.7731} & \textbf{0.7688} & \textbf{0.5656} & \textbf{0.4050} & \textbf{0.3162} \\
\hline
FSIM & 0.5305 & 0.4655 & 0.3271 & 0.5412 & 0.4345 \\
\hline
FSIMc & 0.5292 & 0.4649 & 0.3264 & 0.5417 & 0.4348 \\
\hline
GSI & 0.4248 & 0.3723 & 0.2602 & 0.5780 & 0.4700 \\
\hline
GMSM & 0.7352 & 0.7006 & 0.4996 & 0.4328 & 0.3371 \\
\hline
GMSD & 0.7780 & 0.7345 & 0.5394 & 0.4011 & 0.3094 \\
\hline
LTG & 0.7688 & 0.7309 & 0.5311 & 0.4083 & 0.3155 \\
\hline
VSI & 0.4646 & 0.4104 & 0.2894 & 0.5653 & 0.4568 \\
\hline
ADD-SSIM & 0.6599 & 0.6325 & 0.4626 & 0.4797 & 0.3650 \\
\hline
ADD-GSIM & 0.6521 & 0.6199 & 0.4489 & 0.4841 & 0.3726 \\
\hline
PSIM & 0.7574 & 0.6958 & 0.5010 & 0.4169 & 0.3232 \\
\hline
\end{tabular}}}
\end{table}
\linespread{1}

It can be easily observed that VIFP achieves the best performance in the IPID database compared to other algorithms. Specifically, the PLCC and SROCC values of VIFP exceed 75$\%$. For visualization purposes, we also provide a distribution map of subjective MOS values relative to objective values in the IPID database in Figure 3, where we use "+" to represent distorted images. We can see that the yellow "+" of the VIFP algorithm are uniformly clustered near the diagonal, indicating a better correlation between the scores given by the VIFP algorithm and subjective judgments of image quality. This is because VIFP considers both NSS and HVS to evaluate the perceptual quality of the image. GMSM and PSIM are mainly based on the standard deviation pooling strategy, which significantly improves the efficiency and accuracy of the method and has obvious advantages in various performance indicators. ADD-SSIM also achieves competitive performance by measuring distorted images' contrast and structural information loss. Due to its low correlation with HVS, the performance of SSIM is very poor in this database. By analyzing the experimental results, it can be found that using existing IQA methods to evaluate image quality containing multiple types of distortion remains challenging.
\begin{figure}[!h]
	\small
	\centering
	\includegraphics[width=12cm]{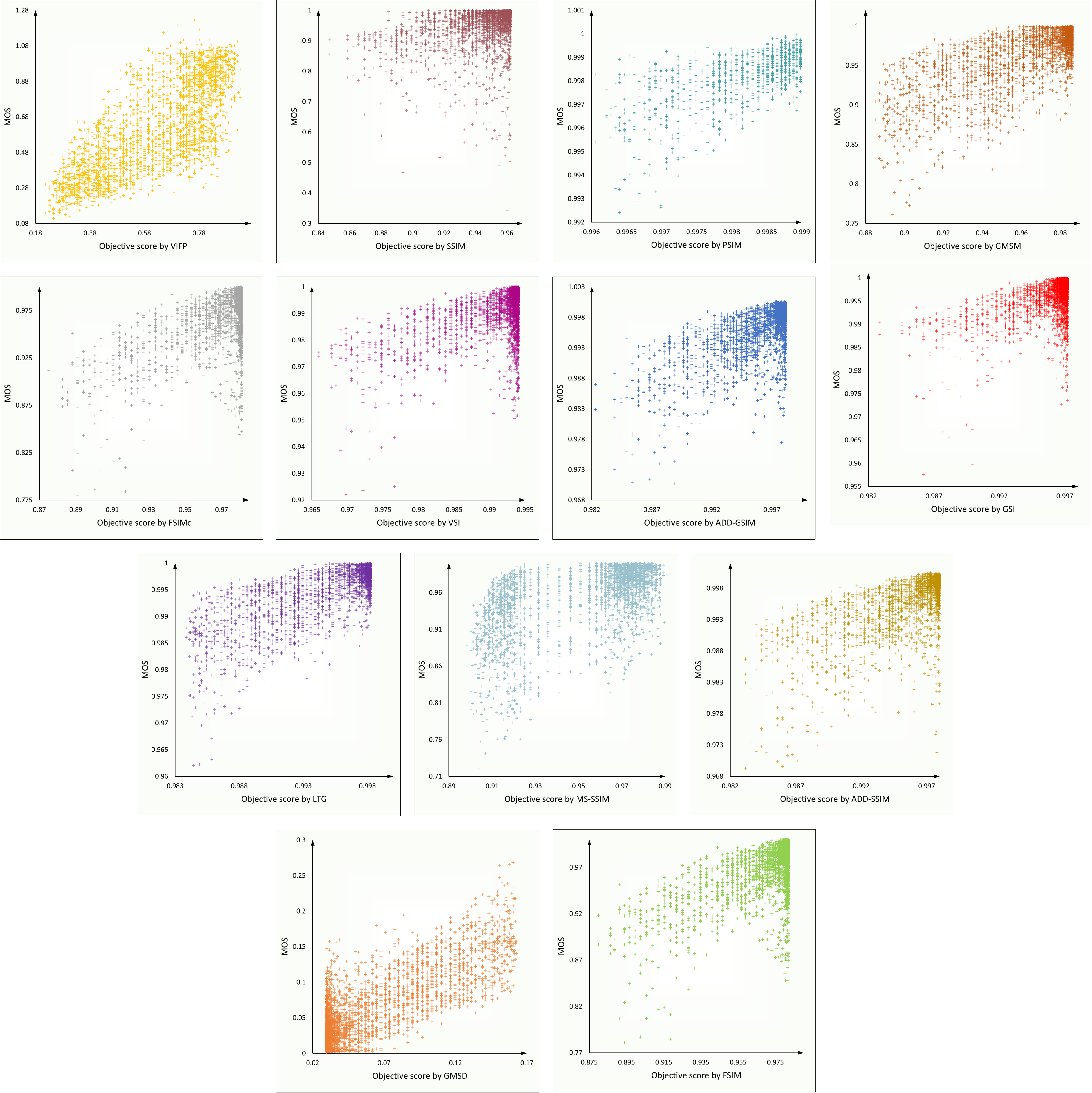}
	\caption{\small   The distribution diagrams of MOS values with respect to objective values on IPID dataset.}
	\label{fig:2}
	\vspace{-0.25cm}
\end{figure}

\section{Conclusion}
In this paper, we firstly establish an industrial process image database called IPID, which consists of 3000 distorted images generated by applying different types and degrees of distortion to 50 source images. Secondly, we conduct subjective scoring experiments on the proposed IPID to obtain image quality assessment scores. Finally, we conduct comparative experiments on the IPID database to investigate the performance of some objective IQA algorithms. The experimental results indicate that most objective IQA algorithms have unsatisfactory correlation with subjective scores and cannot effectively evaluate the quality of distorted images in the database we have established. We will propose a new and effective IQA algorithm to evaluate distorted images in future work.

\end{document}